\documentclass{bmvc2k}

\usepackage[]{amsmath}
\usepackage[]{amssymb}
\usepackage{algorithmic}
\usepackage{algorithm}

\title{Hand-Object Contact Prediction via Motion-Based Pseudo-Labeling and Guided Progressive Label Correction}

\addauthor{Takuma Yagi}{tyagi@iis.u-tokyo.ac.jp}{1}
\addauthor{Md. Tasnimul Hasan}{tasnim@iis.u-tokyo.ac.jp}{1}
\addauthor{Yoichi Sato}{ysato@iis.u-tokyo.ac.jp}{1}

\addinstitution{
 The University of Tokyo\\
 Tokyo, Japan
}

\runninghead{Yagi, Hasan, Sato}{Hand-Object Contact Prediction}

\def\eg{\emph{e.g}\bmvaOneDot}

\def\etal{\emph{et al}\bmvaOneDot}

\def\eg{{\it e.g.}}

\begin{document}

\maketitle

\begin{abstract}
Every hand-object interaction begins with contact. 
Despite predicting the contact state between hands and objects is useful in understanding hand-object interactions, prior methods on hand-object analysis have assumed that the interacting hands and objects are known, and were not studied in detail.
In this study, we introduce a video-based method for predicting contact between a hand and an object.
Specifically, given a video and a pair of hand and object tracks, we predict a binary contact state (contact or no-contact) for each frame.
However, annotating a large number of hand-object tracks and contact labels is costly.
To overcome the difficulty, we propose a semi-supervised framework consisting of (i) automatic collection of training data with motion-based pseudo-labels and (ii) guided progressive label correction (gPLC), which corrects noisy pseudo-labels with a small amount of trusted data.
We validated our framework's effectiveness on a newly built benchmark dataset for hand-object contact prediction and showed superior performance against existing baseline methods.
Code and data are available at \url{https://github.com/takumayagi/hand_object_contact_prediction}.
\end{abstract}

\section{Introduction}

\begin{figure}[t]
\centerline{\includegraphics[width=0.9\linewidth]{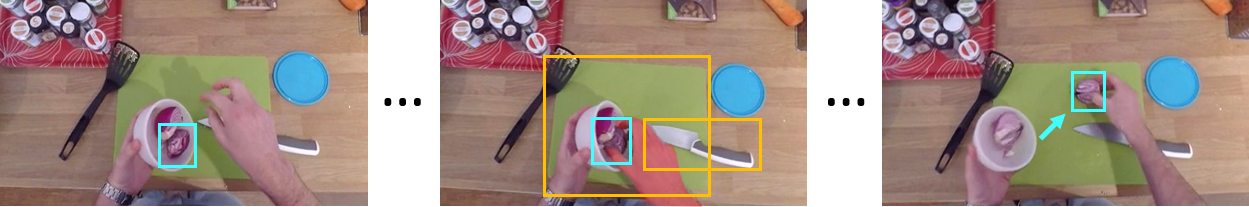}}
\caption{{\bf Overlap on Image = Contact?:} Right hand (masked red region) grabs the onion (middle). Surrounding objects (cutting board, knife) overlaps with hand but not in contact. While it is difficult to determine contact state from single image, we can ease the problem by looking at temporal context (left and right).}
\label{fig:teaser}
\vspace{-1.5em}
\end{figure}

Recognizing how hands interact with objects is crucial to understand how we interact with the world.
Hand-object interaction analysis contributes to several fields such as action prediction~\cite{dessalene2021forecasting}, rehabilitation~\cite{likitlersuang2019egocentric}, robotics~\cite{rajeswaran2018learning}, and virtual reality~\cite{han2020megatrack}.

Every hand-object interaction begins with contact.
In determining which hand-object pairs are interacting, it is important to infer when hands and objects are in contact.
However, despite its importance, finding the beginning and the end of hand-object interaction has not received much attention. For instance, prior works on action recognition (\eg, \cite{Feichtenhofer_2019_ICCV}) attempt to recognize different types of hand object interactions at the video clip level, i.e., recognizing one action for each video clip given as input. Some other works on action localization (\eg, \cite{Lin2019BMNBN}) can be used for detecting hand object interactions but localized action segments are not necessarily related to the beginning and the end of contact between hands and objects. 
Contact between a hand and an object has been studied in the context of 3D reconstruction of hand object interaction~\cite{hasson2019learning,cao2020reconstructing}.
However, they assumed that hands and objects are already interacting with each other.
Only the moment when hands and objects are interacting in a stable grasp was targeted for analysis.

In this work, we tackle the task of predicting contact between a hand and an object from visual input. Predicting contact between a hand and an object from visual input is not trivial.
For example, even if the hand area and the bounding box of an object overlap, it does not necessarily mean that the hand and the object are in contact (see Figure~\ref{fig:teaser}).
In determining whether a hand and an object are in contact, it is essential to consider the spatiotemporal relationship between them.
While some methods claim that the hand contact state can be classified by looking at hand shape~\cite{shan2020understanding,narasimhaswamy2020detecting}, they did not explicitly predict the contact state between a specific pair of a hand and an object, limiting their utility.

We propose a video-based method for predicting binary contact states (contact or no-contact) between a hand and an object in every frame.
We assume tracks of hands and objects specified by bounding boxes ({\it hand-object tracks}) as input.
However, annotating a large number of hand-object tracks and their contact states can become too costly.
To overcome this difficulty, we propose a semi-supervised framework consisting of (i) automatic training data collection with motion-based pseudo-labels and (ii) guided progressive label correction (gPLC) which corrects noisy pseudo-labels with a small amount of trusted data.

Given unlabeled videos, we apply off-the-shelf detection and tracking models to form a set of hand-object tracks.
Then we assign pseudo-contact state labels to each track by looking at its motion pattern.
Specifically, we assign a contact label when a hand and an object are moving in the same direction and a no-contact label when a hand is moving alone.

While generated pseudo-labels can provide valuable information on determining the state of contact states with various types of objects when training a prediction model, the pseudo-labels also contain errors that hurt the model's performance.
To alleviate this problem, we correct those errors by the guidance of an additional model trained on a small amount of trusted data.
In gPLC, we train two networks each trained with noisy labels and trusted labels.
During the training, we iteratively correct noisy pseudo-labels based on both network's confidence scores.
We use the small-scale trusted data to guide which label to be corrected and yield reliable training labels for automatically extracted hand-object tracks.

Since there was no benchmark suitable for this task, we newly annotated contact states to various types of interactions appearing in the EPIC-KITCHENS dataset~\cite{damen2018scaling,Damen2020RESCALING} which includes in-the-wild cooking activities.
We show that our prediction model achieves superior performance against frame-based models~\cite{shan2020understanding,narasimhaswamy2020detecting}, and the performance further boosted by using motion-based pseudo-labels along with the proposed gPLC scheme.

Our contributions include: (1) A video-based method of predicting contact between a hand and an object leveraging temporal context; (2) A semi-supervised framework of automatic pseudo-contact state label collection and guided label correction to complement lack of annotations; (3) Evaluation on newly collected annotation over a real-world dataset.

\section{Related Works}

\paragraph*{Reconstructing hand-object interaction}
Reconstruction of the spatial configuration of hands and their interacting objects plays a crucial role to understand hand-object interaction.
2D segmentation~\cite{shilkrot2019workinghands} and 3D pose/mesh estimation~\cite{romero2010hands,tzionas20153d,hasson2019learning,hampali2020honnotate,liu2021semi} of hand-object interaction were studied actively in recent years.
However, they assume (1) 3D CAD models exist for initialization (except{~\cite{hasson2019learning}}) (2) the hand is interacting with objects, making the methods inapplicable when hand and object are not interacting with each other.
While multiple datasets appear for hand-object interaction analysis~\cite{garcia2018first,shilkrot2019workinghands,brahmbhatt2019contactdb,shan2020understanding}, no dataset focused on the entire process of interaction including beginning and termination of contact.
It is worth mentioning DexYCB~{\cite{chao2021dexycb}}, which captured sequences of picking up an object.
However, the performed action was very simple and their analysis focused on 3D pose estimation rather than contact modeling between hands and objects.
We study the front stage of the hand-object reconstruction problem---whether the hand interacts with the object or not.

\paragraph*{Hand-object contact prediction}
Contact prediction is found to be a difficult problem because contact cannot be directly observed due to occlusions.
To avoid using intrusive hand-mounted sensors, contact and force prediction from visual input was studied~\cite{pham2017hand,akizuki2018tactile,taheri2020grab,ehsani2020use}.
For example, Pham \etal~\cite{pham2017hand} present an RNN-based force estimation method trained on force and kinematics measurements from force transducers.
These methods require a careful setup of sensors, making it hard to apply them in an unconstrained environment.
Instead of precise force measurement, a few methods study contact state classification (\eg, no contact, self contact, other people contact, object contact) from an image~\cite{shan2020understanding,narasimhaswamy2020detecting}.
Shan \etal~\cite{shan2020understanding} collected a large-scale dataset of hand-object interaction along with annotated bounding boxes of hands and objects in contact.
They train a network that detects hands and their contact state from their appearance.
Narasimhaswamy \etal~\cite{narasimhaswamy2020detecting} extends the task into multi-class prediction.
While their formulation is simple, they did not take the relationship between hands and objects explicitly and were prone to false-positive prediction.
To balance utility and convenience, we take the middle way between the two approaches---binary contact state prediction between a hand and an object specified by bounding boxes.

\paragraph*{Learning from noisy labels}
Since dense labels are often costly to collect, methods to learn from large unlabeled data are studied.
While learning features from weak cues are studied in object recognition~\cite{pathak2017learning} and instance segmentation~\cite{pathakCVPRW18learning}, it was not well studied in a sequence prediction task.
Generated pseudo-labels typically include noise that hurts the model's performance.
Various approaches such as loss correction~\cite{patrini2017making,hendrycks2018using}, label correction~\cite{tanaka2018joint,zhang2021learning}, sample selection~\cite{jiang2018mentornet}, and co-teaching~\cite{han2018co,li2020dividemix} are proposed to deal with noisy labels.
However, most methods assume feature-independent feature noise which is over-simplified, and only a few works study realistic feature-dependent label noise~\cite{chen2020beyond,zhang2021learning}.
Zhang \etal~\cite{zhang2021learning} propose progressive label correction (PLC) which iteratively corrects labels based on the network's confidence score with theoretical guarantees against feature-dependent noise patterns.
Inspired by PLC~\cite{zhang2021learning}, we propose gPLC which iteratively corrects noisy labels by not only the prediction model but also with the clean model trained on small-scale trusted labels.

\section{Proposed Method}

In contrast to prior works~\cite{shan2020understanding,narasimhaswamy2020detecting} we formalize the hand-object contact prediction problem as predicting the contact states between a hand and a specific object appearing in a image sequence.
We assume video frames $\mathcal{X}=\{X^1, \dots, X^T\}$, hand instance masks $\mathcal{H}=\{H^1, \dots, H^T\}$, and target object bounding boxes $\mathcal{O}=\{O^1, \dots, O^T\}$ as inputs, forming a hand-object track $\mathcal{T}=(\mathcal{X}, \mathcal{H}, \mathcal{O})$.

Our goal is to predict a sequence of a binary contact state (``no contact'' or ``contact'') ${\bf y}=\{y^1, \dots, y^T\} (y\in \{0, 1\})$ given a hand-object track $\mathcal{T}$.
If any physical contact between the hand and the object exists, the binary contact state $y$ is set to $1$, otherwise $0$.
Although we do not explicitly model two-hands manipulation, we consider the presence of another hand as side information (see Section {\ref{ssec:contact_prediction}} for details).

However, collecting a large number of hand-object tracks and contact states for training can become too costly.
We deal with this problem by automatic pseudo-label collection based on motion analysis and a semi-supervised label correction scheme.

\subsection{Pseudo-Label Generation from Motion Cues}
\begin{figure}[t]
\centerline{\includegraphics[width=0.9\linewidth]{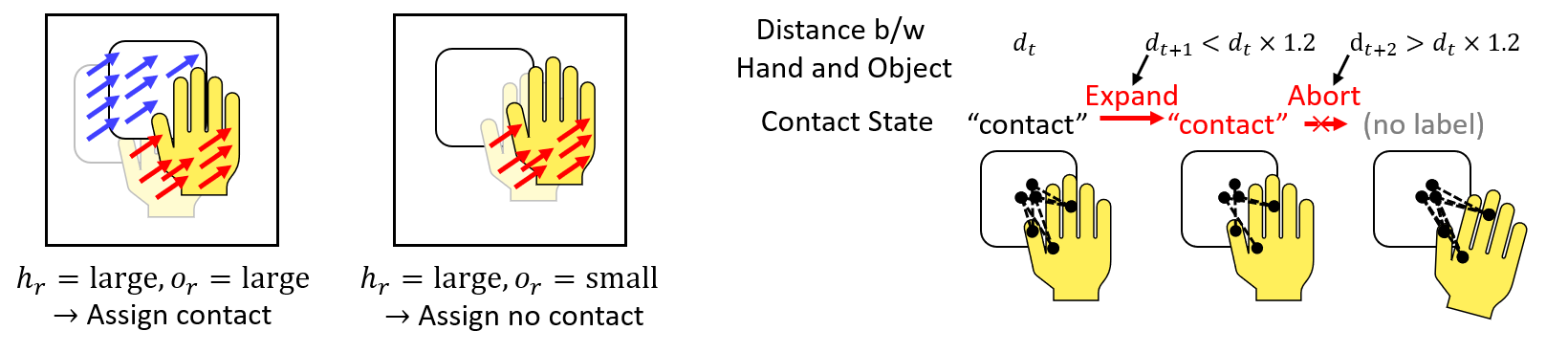}}
\caption{(Left) Pseudo-label generation from motion cues. (Right) Pseudo-label extension based on hand-object distance.}
\label{fig:expansion}
\vspace{-1em}
\end{figure}

We automatically detect hand-object tracks and assign pseudo-labels to them based on two critical assumptions. (i) When a hand and an object are in contact, they exhibit similar motion pattern. (ii) When a hand and an object are not in contact, the hand moves while the object remains static (see Figure~\ref{fig:expansion} left for illustration).
Because these assumptions are simple yet applicable regardless of object appearance and motion direction, we can use these motion-based pseudo-labels for training to achieve generalization against novel objects.

\paragraph*{Hand-object track generation}
Given a video clip, we first use the hand-object detection model~\cite{shan2020understanding} to detect bounding boxes of hands and candidate objects appearing in each frame.
Note that the detected object's contact state is unknown and objects which overlap with hands are detected.
For each hand detection, we further apply a hand segmentation model trained on EGTEA dataset~\cite{li2018eye} to each hand detection to obtain segmentation masks.

Next, we associate adjacent detections using a Kalman Filter-based tracker~\cite{bewley2016sort}.
However, since \cite{shan2020understanding} does not detect objects away from the hand, we extrapolate object tracks one second before and after using a visual tracker~\cite{li2019siamrpn++}, producing $\mathcal{H}$ and $\mathcal{O}$.
Finally, we construct the hand-object track $\mathcal{T}$ by looking for pairs of hand and object tracks which include a spatial overlap between hand mask and object bounding box.

\paragraph*{Contact state assignment}
We find contact (and no-contact) moments by looking at the correlation between hand and object motion.
First, we estimate optical flows between adjacent frames.
Since we are interested in relative movement of hands and objects against backgrounds, we obtain background motion-compensated optical flow and its magnitude $M$ by homography estimation.
Specifically, we sample flow vectors outside detected bounding boxes as matches between frames and estimate the homography using RANSAC~\cite{fischler1981random}.

Let $F=(f_{ij})=\mathbb{I}_{(M > \sigma)}$ be a binary mask of foreground moving region its magnitude larger than a certain threshold $\sigma$.
For each hand and object binary region mask $H=(h_{ij})$ and $O=(o_{ij})$, we calculate the ratio of moving region within each region: $h_r=\frac{\sum_{ij} (h_{ij} \cdot f_{ij})}{\sum_{ij} h_{ij}}$, $o_r=\frac{\sum_{ij} (o_{ij} \cdot f_{ij})}{\sum_{ij} o_{ij}}$.
We assign a label to a frame if $\mathrm{IoU}(H, O) > 0$ and $h_r$ and $o_r$ above certain thresholds.
Similarly, we assign a no-contact label if $\mathrm{IoU}(H, O) = 0$ or $h_r$ above threshold but $o_r$ below threshold.
However, the above procedure may wrongly assign contact labels if the motion direction of hand and object are different (\eg, the object handled by the other hand).
Thus we calculate the cosine similarity between the average motion vector of hand and object region and assign a contact label if above threshold otherwise a no-contact label.
To deal with errors in flow estimation, we cancel the assignment if the background motion ratio $b_r=\frac{\sum_{ij} (b_{ij} \cdot f_{ij})}{\sum_{ij} b_{ij}}$ ($B=(b_{ij})$ denotes background mask other than $H$ and $O$) is above threshold.

\paragraph*{Pseudo-label extension}

\begin{figure}[t]
\centerline{\includegraphics[width=0.9\linewidth]{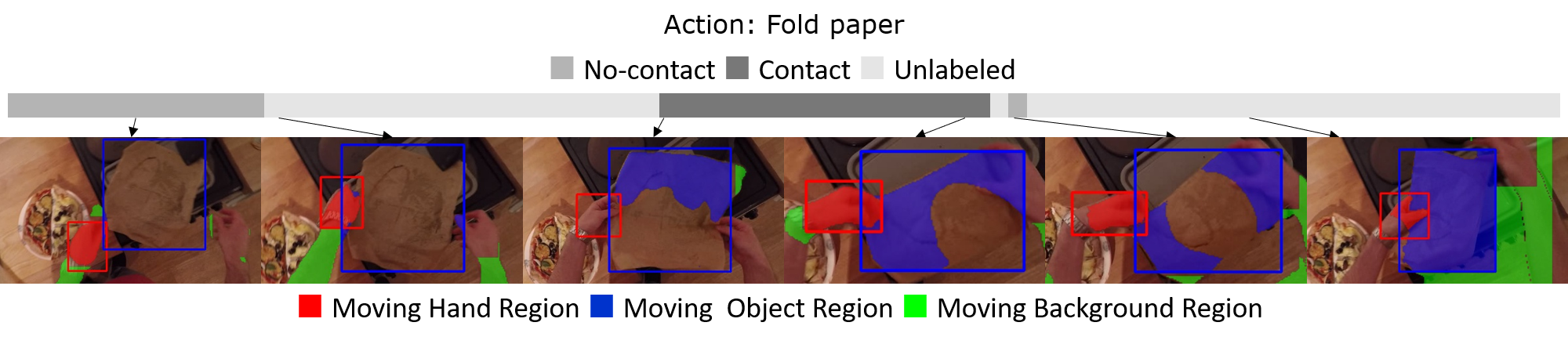}}
\caption{{\bf Example of generated pseudo-labels:} (Top) Gray and dark gray bar indicates no-contact/contact labels otherwise no labels assigned. (Bottom) Representative frames. Red, blue, and green regions denote moving hand, object, and background regions, respectively. In rightmost frame, no label is assigned because of abrupt background motion.}
\label{fig:pseudo_labels}
\vspace{-1em}
\end{figure}

The above procedure assigns labels on hand-moving frames, but it does not assign labels when hands are moving slowly or still.
To assign labels also on those frames, we extend the assigned contact states if the relationship between hands and objects does not change from the timing when pseudo-labels are assigned (see Figure~\ref{fig:expansion} right).

To track hand-object distance, we find point trajectories from hand and object region which satisfy forward-backward consistency~\cite{sundaram2010dense}.
We then calculate the distance $d$ between each hand-object point pair and compare the average distance of them in each frame.
We extend the last contact state if the average distance is within a certain range of that of the starting frame.
Figure~\ref{fig:pseudo_labels} shows an example of the generated pseudo-labels.

\subsection{Guided Progressive Label Correction (gPLC)}
\begin{figure}[t]
\begin{tabular}{c}
\begin{minipage}{0.4\hsize}
\centerline{\includegraphics[width=1.0\linewidth]{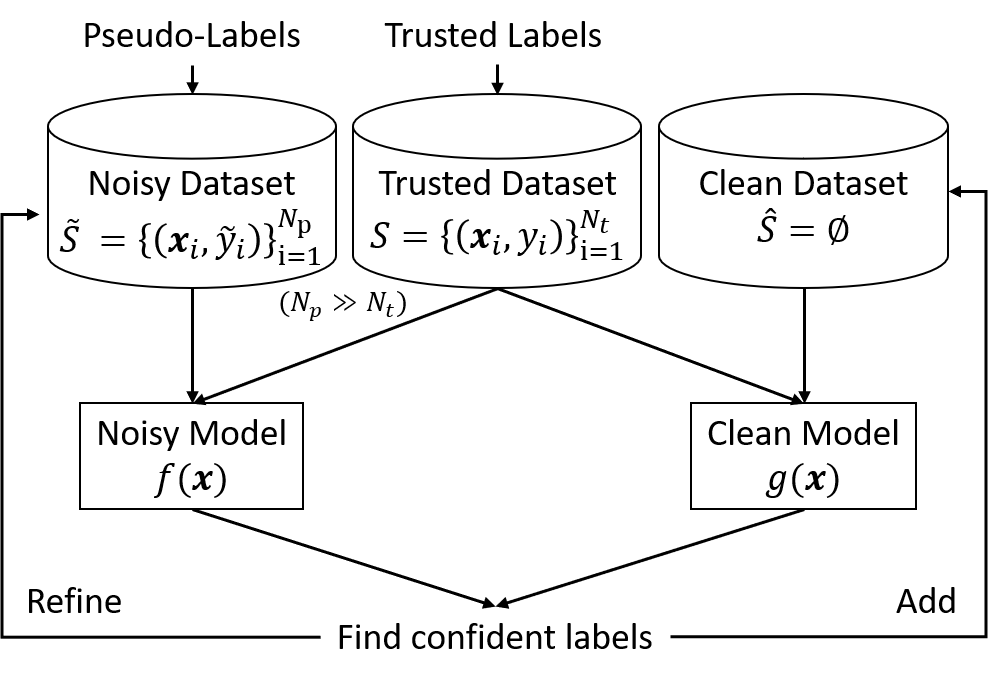}}
\caption{Overview of guided progressive label correction (gPLC).}
\label{fig:gplc}
\end{minipage}
\begin{minipage}{0.6\hsize}
\centerline{\includegraphics[width=0.95\linewidth]{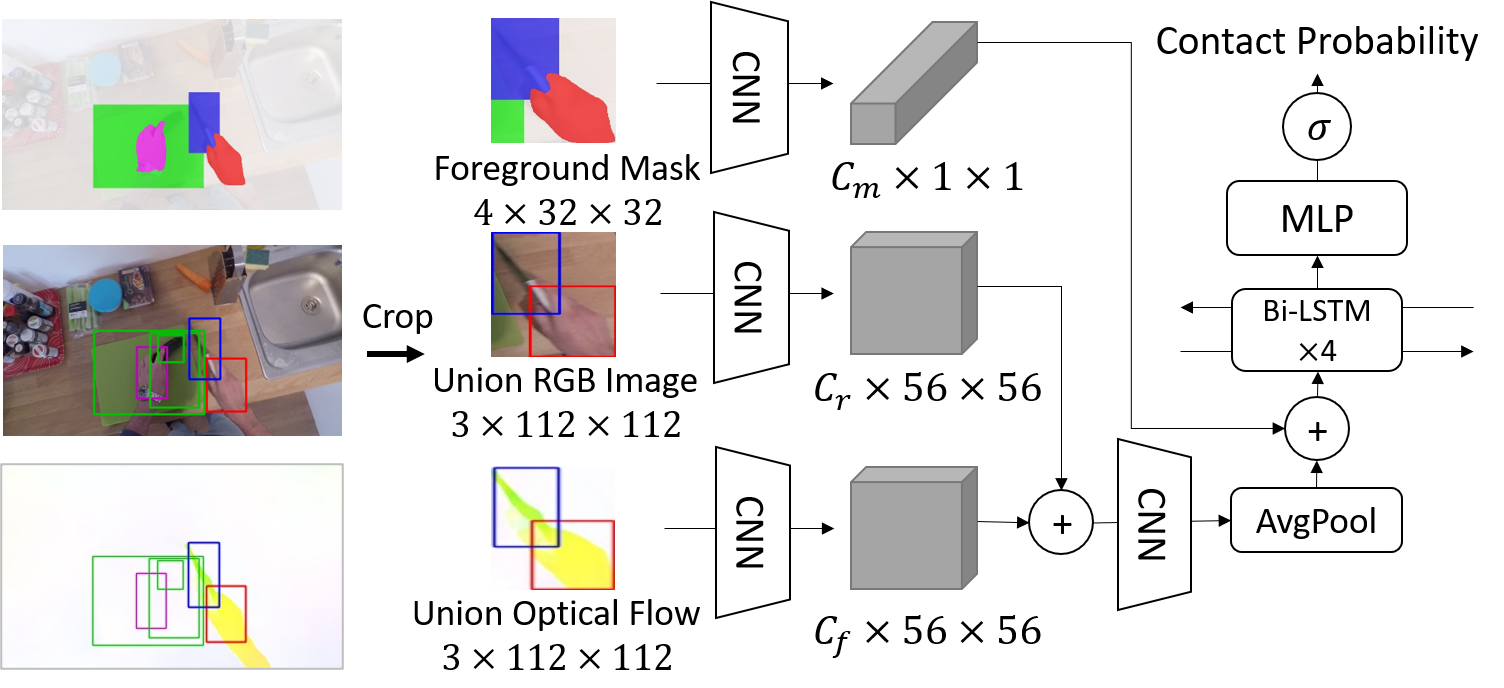}}
\caption{Architecture of contact prediction model.}
\label{fig:method}
\end{minipage}
\end{tabular}
\vspace{-1em}
\end{figure}

\begin{figure}[t]
\begin{center}
\scalebox{0.9}{
\begin{minipage}{1.0\linewidth}
\begin{algorithm}[H]
    \algsetup{linenosize=\small}
    \small
    \caption{Guided Progressive Label Correction (gPLC) }
    \label{alg:gplc}
    \begin{algorithmic}
    \REQUIRE Noisy dataset $\tilde{S}=\{({\bf x}_i, \tilde{{\bf y}}_i)\}_{i=1}^{N_p}$, trusted dataset $S=\{({\bf x}_i, {\bf y}_i)\}_{i=1}^{N_t}$, clean dataset $\hat{S}=\varnothing$, noisy model $f({\bf x})$, clean model $g({\bf x})$, initial and end thresholds $(\delta_0, \delta_{end})$, correction threshold $\delta=\delta_0$, flip ratio $\alpha$, step size $\beta$,  supervision interval $m$, total round $N$
    \ENSURE Trained Model $f({\bf x})$
    \STATE $\hat{S}=\{({\bf x}_i, \hat{{\bf y}}_i)\}_{i=1}^{N_p}$ where $\hat{{\bf y}_i}$ is a list with empty elements same size as $\tilde{{\bf y}}_i$ \hfill// Initialize clean dataset
    \FOR{$n \leftarrow 1,\dots,N$}
        \FOR{$i \leftarrow 1,\dots,N_p$}
            \STATE ${\bf z}_i \leftarrow {\bf \tilde{y}}_i$ \hfill// Keep previous labels
            \FOR{$t \leftarrow 1,\dots, |\tilde{{\bf y}}_i|$}
                \IF{$\tilde{y}_i^t \in \{0, 1\}$ \AND $|f({\bf x}_i^t) - \frac{1}{2}| \ge \frac{1}{2} - \delta$ \AND $\mathbb{I}_{\{f({\bf x}_i^t) \ge \frac{1}{2}\}} = \mathbb{I}_{\{g({\bf x}_i^t) \ge \frac{1}{2}\}}$}
                    \STATE $\tilde{y}_i^t, \hat{y}_i^t \leftarrow \mathbb{I}_{\{f({\bf x}_i^t) \ge \frac{1}{2}\}}$ \hfill// Refine or add label by confident prediction
                \ENDIF
            \ENDFOR
            \STATE Train $f({\bf x})$ on $({\bf x}_i, \tilde{{\bf y}}_i)$ and $g({\bf x})$ on $({\bf x}_i, \hat{{\bf y}}_i)$ \hfill// Update models 
            \IF{\#iterations \% $m$ = 0}
                \STATE Train $f({\bf x})$ and $g({\bf x})$ on $S$ \hfill // Fine-tune on trusted set
            \ENDIF
        \ENDFOR
        \IF{$\sum_{i, t} \mathbb{I}_{\{\tilde{y}_i^t = z_i^t\}}$ < $\alpha \cdot \sum_{i, t} \mathbb{I}_{\{\tilde{y}_i^t \in \{0, 1\}\}}$}
            \STATE $\delta \leftarrow min(\delta + \beta, \delta_{end})$ \hfill// Loosen threshold if number of flipped labels are small enough
        \ENDIF
    \ENDFOR
    \end{algorithmic}
\end{algorithm}
\end{minipage}
}
\end{center}
\vspace{-2em}
\end{figure}

While generated pseudo-labels include useful information in determining contact states, they also include errors induced by irregular motion patterns.
The model may overfit to noise if we simply train it based on these noisy labels.
To utilize reliable labels from noisy pseudo-labels, we propose a semi-supervised procedure called guided progressive label correction (gPLC), which works with a small number of trusted labels (see Figure~\ref{fig:gplc} for overview).

We assume a noisy dataset $\tilde{S}$ with generated pseudo-labels and a trusted dataset $S$ with manually annotated trusted labels.
We train two identical networks, each called noisy model and clean model.
The noisy model $f$ is trained on both $\tilde{S}$ and $S$ while the clean model $g$ is trained on $S$ and a clean dataset $\hat{S}$ which is introduced later.
We perform label correction against generated pseudo-labels in $\tilde{S}$ using the prediction of both models.

As the training of $f$ proceeds, $f$ will give high confidence against some samples.
Similar to PLC~\cite{zhang2021learning}, we try to correct labels on which $f$ gives high confidence.
Note that we correct labels in a frame-wise manner, assuming output contact probability is produced per frame.
In gPLC, we correct labels only when $f$ has high confidence and does not contradict the clean network $g$'s prediction.
Because $\tilde{S}$ is generated from motion cues, the decision boundary of $f$ may be different from that of the optimal classifier.
Thus the label correction on $f$ alone would not converge to the desired decision boundary.
Therefore, we guide the correction process by using $g$. %
Starting with a strict threshold on $\delta$, we iteratively correct labels upon training.
When the number of corrected labels gets small enough, we increase $\delta$ to loosen the threshold and continue the same procedure.
However, since $g$ is trained on a small-scale data, it has the risk of overfitting to $S$.
To prevent overfitting, we iteratively add data that $f$ gives high confidence to another dataset called clean dataset $\hat{S}$ and feed them to $g$ so that $g$ also grows through training.
Initially $\hat{S}$ will not contain labels, but high-confident labels will be added over time.
See Algorithm~{\ref{alg:gplc}} for detail.
In implementation, $f(\mathbf{x})$ and $g(\mathbf{x})$ are trained beforehand by $\tilde{S}$ and $S$ before starting the gPLC iterations.

\subsection{Contact Prediction Model}
\label{ssec:contact_prediction}

Contact states are predicted by using an RNN-based model that takes RGB images, optical flow, and mask information as input (see Figure~\ref{fig:method}).
For each modality, we crop the input by taking the union of the hand region and the object bounding box.
The foreground mask is a four-channel binary mask that tells the presence of a target hand instance mask, a target object bounding box, other detected hand instance masks, and other detected object bounding boxes.
The latter two channels prevent confusion when the target hand or object interacts with other entities.
RGB and flow images are fed into each encoder branch, concatenated at the middle, and then passed to another encoder.
Both encoders consist of several convolutional blocks, each consisted of 3$\times$3 convolution followed by a ReLU and a LayerNorm layer~\cite{ba2016layer}, and a 2$\times$2 max-pooling layer. %
The foreground mask encoder consists of three convolutional layers each followed by a ReLU layer, producing a 1$\times$1 feature map encoding the positional relationship between the target hand, the target object, and the other hands and objects.
After concatenating the features extracted from the foreground mask, contact probability is calculated through four bi-directional LSTM layers and three layers of MLP.

We train the network by a standard binary cross-entropy loss weighted by the ratio of the amount of labels in the training data.
We did not propagate the error for non-labeled frames.

\section{Experiments}
Since there was no benchmark suitable for our task, we newly annotated hand-object tracks and contact states between hands and objects against videos in EPIC-KITCHENS dataset~\cite{damen2018scaling}.
We collected tracks with various objects (\eg, container, pan, knife, sink).
The amount of the annotation was 1,200 tracks (67,000 frames) in total.
We split the data into a training set (240 tracks), validation set (260 tracks), and test set (700 tracks).
For the noisy dataset, we have generated 96,000 tracks with motion-based pseudo-labels.

\subsection{Implementation Details}
We used FlowNet2~\cite{ilg2017flownet} for optical flow estimation. We used Adam~\cite{kingma2014adam} for optimization with a learning rate of 3e-4. We trained the network for 500,000 iterations with a batch size of one and selected the best model by frame accuracy on the validation set.
The hyperparameters were set to $\delta_0=0.05, \delta_{end}=0.25, \alpha=0.01, \beta=0.025, m=2500$.
\subsection{Evaluation Metrics}
We prepared several metrics to evaluate the performance.
{\bf Frame Accuracy: } Frame-wise accuracy balanced by the ground truth label ratio;
{\bf Boundary Score: } F-measure of boundary detection. Performs bipartite graph matching between ground truth and predicted boundary~\cite{perazzi2016benchmark}. Count as correct if the predicted boundary within six frames from the ground truth boundary;
{\bf Peripheral Accuracy: } Frame-wise accuracy within six frames from the ground truth boundary;
{\bf Edit Score: } Segmental metric using Levenshtein distance between segments~\cite{lea2016segmental}. We assume both contact and no-contact labels are foreground;
{\bf Correct Track Ratio: } The ratio of tracks which gives frame accuracy above 0.9 and boundary score of 1.0.

\subsection{Baseline Methods}
We evaluated several baseline methods:
{\bf Fixed:} Predicts always as ``contact'';
{\bf IoU:} Calculate the mask IoU between the input hand mask and object bounding box. If the score is larger than zero predicts as contact, otherwise no-contact;
{\bf ContactHands~\cite{narasimhaswamy2020detecting}:} Predicts as a contact if the detected hand's contact state is ``object'';
{\bf Shan-Contact~\cite{shan2020understanding}:} Predicts as a contact if corresponding hand's contact state prediction is ``portable'';
{\bf Shan-Bbox~\cite{shan2020understanding}:} Predicts as contact if there is enough overlap between the detected object bounding box and input object bounding box;
{\bf Shan-Full~\cite{shan2020understanding}:} Combines predictions of Shan-Contact and Shan-Bbox;
{\bf Supervised:} Our proposed prediction model, trained by trusted data alone.
We note that for the \textbf{Shan-\textasteriskcentered} baselines, the {\it 100k+ego} pre-trained model provided by the authors was used, which is trained on egocentric video datasets including the EPIC-KITCHENS dataset.

\subsection{Results}
\paragraph*{Quantitative results}
\begin{table}[t]
\begin{center}
\scalebox{.75}{
\begin{tabular}{lrrrrr}
\hline
Method & Frame Acc. & Boundary Score & Peripheral Acc. & Edit Score & Correct Ratio\\ \hline
Fixed & 0.500 & 0.394 & 0.534 & 0.429 & 0.166 \\
IoU & 0.642 & 0.505 & 0.613 & 0.678 & 0.259 \\ \hline
ContactHands~\cite{narasimhaswamy2020detecting} & 0.555 & 0.440 & 0.596 & 0.468 &  0.136 \\
Shan-Contact~\cite{shan2020understanding} & 0.608 & 0.516 & 0.656 & 0.507 & 0.180 \\
Shan-Bbox~\cite{shan2020understanding} & 0.688 & 0.435 & 0.639 & 0.631 &  0.189 \\
Shan-Full~\cite{shan2020understanding} & 0.746 & 0.477 & 0.687 & 0.583 &  0.193 \\ \hline
Supervised (train) & 0.770 & 0.563 & 0.649 & 0.718 & 0.394 \\ %
Supervised (train+val) & 0.816 & 0.636 & 0.695 & {\bf 0.793} & 0.487 \\ %
Proposed & {\bf 0.836} & {\bf 0.681} & {\bf 0.730} & {\bf 0.793} & {\bf 0.519} \\ \hline %
\end{tabular}
}
\caption{Results of hand contact state prediction performance.}
\label{tab:result}
\end{center}
\vspace{-1em}
\end{table}

\begin{table}[t]
\begin{center}
\scalebox{.73}{
\begin{tabular}{lrrrrr}
\hline
Method & Frame Acc. & Boundary Score & Peripheral Acc. & Edit Score & Correct Ratio\\ \hline
Noisy Label only & 0.780 & 0.569 & 0.703 & 0.687 & 0.344 \\ %
Noisy + Trusted Label & 0.811 & 0.624 & 0.708 & 0.759 & 0.453 \\ \hline %
Noisy + Trusted Label w/ PLC~\cite{zhang2021learning} & 0.821 & 0.636 & 0.730 & 0.768 & 0.480 \\ \hline %
Pseudo-Labeling~\cite{lee2013pseudo} & 0.784 & 0.590 & 0.703 & 0.737 & 0.417 \\ \hline %
RGB & 0.787 & 0.546 & 0.681 & 0.709 & 0.363 \\ %
Flow & 0.833 & 0.672 & 0.725 & 0.789 & 0.519 \\ \hline %
Proposed (RGB+Flow) & 0.836 & 0.681 & 0.730 & 0.793 & 0.519 \\ \hline %
\end{tabular}
}
\caption{Ablations on input modality and other robust learning methods.}
\label{tab:ablation_2}
\end{center}
\vspace{-1em}
\end{table}

We report the performance in Table~\ref{tab:result}.
Our proposed method consistently outperforms the baseline models on all the metrics, achieving a double correct track ratio compared to {\bf IoU} based on the overlap between hand and object bounding boxes.
The frame-based methods ({\bf ContactHands}, {\bf Shan-\textasteriskcentered}) performed equal or worse than {\bf IoU}, producing many false positive predictions.
These results suggest that previous methods claiming contact state prediction fails to infer physical contact between hands and objects.
While {\bf Supervised} performed well, gPLC further boosted the performance by leveraging diverse motion-based cues with label correction, especially on boundary score.

\paragraph*{Qualitative results}

\begin{figure}[t]
\begin{tabular}{c}
    \begin{minipage}[t]{1\hsize}
    \centerline{\includegraphics[width=0.85\linewidth]{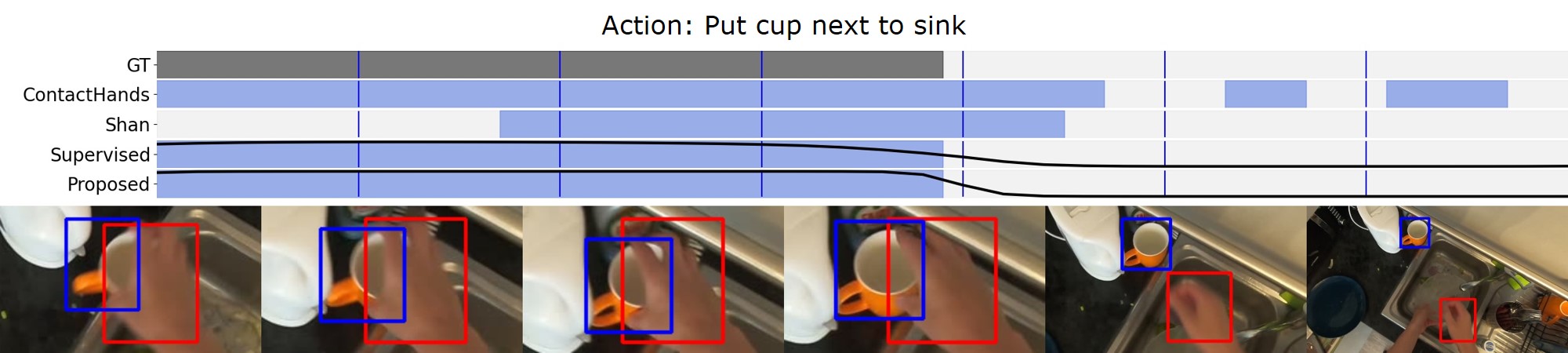}} %
    \end{minipage} \\ 
    \begin{minipage}[t]{1\hsize}
    \centerline{\includegraphics[width=0.85\linewidth]{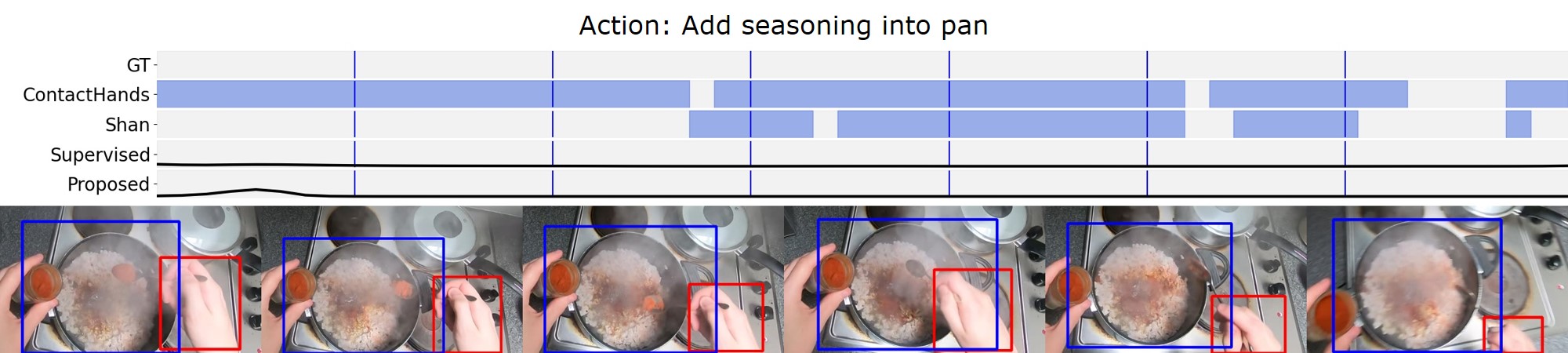}} %
    \end{minipage} \\
    \begin{minipage}[t]{1\hsize}
    \centerline{\includegraphics[width=0.85\linewidth]{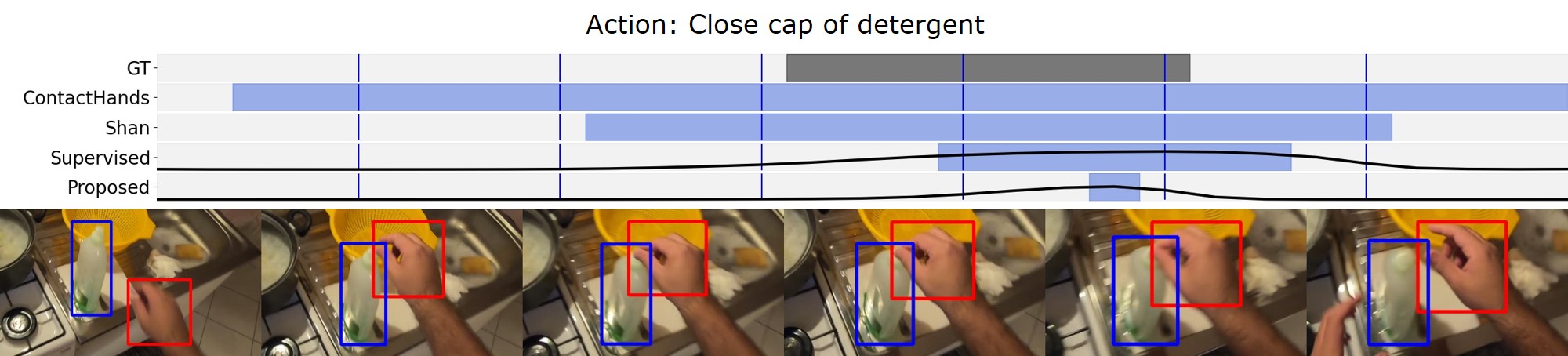}} %
    \end{minipage}
\end{tabular}
\caption{Qualitative examples. Upper chart shows ground truth contact state and prediction of each model (gray and blue region indicates contact, otherwise no-contact) with contact probability in black line. Lower images correspond to blue vertical lines in chart from left to right and red and blue boxes represents input hand and object bounding box.}
\label{fig:qual_result}
\vspace{-1em}
\end{figure}

Figure~\ref{fig:qual_result} shows the qualitative results.
As shown in the top, our method distinguish contact and no-contact states by looking at the interaction between hands and objects while baseline methods yield false positive predictions by looking at box overlaps.
The middle shows a typical no-contact case of a hand floating above an object.
Our proposed model trained on motion-based pseudo-labels avoid producing false positive prediction.

\paragraph*{Comparison against other robust learning methods}

To show the effectiveness of the proposed gPLC, we report ablations on other robust learning/semi-supervised learning methods (see Table~\ref{tab:ablation_2} top).
As expected, training using motion-based pseudo-labels performed worse due to labeling errors.
Joint training with noisy and trusted labels gives marginal gain against the supervised model, but the boundary score remains low since it overfits against pseudo-label noise.
We also applied the existing label correction method~{\cite{zhang2021learning}} on a single network with fine-tuning on trusted labels, but its performance was almost equal to joint training, suggesting that label correction on a single network does not yield good correction.
We also tried a typical pseudo-labeling~\cite{lee2013pseudo} without motion-based labels.
However, it showed only a marginal improvement over the supervised baseline, suggesting that our motion-based pseudo-labels are necessary for better generalization.

\paragraph*{Effect of input modality}

The bottom of Table~\ref{tab:ablation_2} reports the ablation results of changing the input modalities.
We observed that using RGB images alone impacts the boundary score, suggesting the difficulty of determining the contact state change without motion information.
In contrast, the optical flow-based model achieved nearly the same performance as the full model, suggesting that motion information is crucial for accurate prediction.

\paragraph*{Error analysis}
While our method can better predict contact states by utilizing the rich supervision from motion-based pseudo-labels, we observed several failure patterns.
As shown in Figure~\ref{fig:qual_result} bottom, our method often ignored contacts when a person instantly touched objects without yielding apparent object motion.
We also observed failures due to unfamiliar grasps, complex in-hand motion, and failure in determining object regions (see supplemental for more results).
These errors indicate the limitation of the motion-based pseudo-labels which assigns labels only when clear joint motion is observed.
To better deal with subtle/complex hand motions, additional supervision or rules on such patterns may be required.

\paragraph*{How does gPLC correct noisy labels?}
\begin{figure}[t]
\begin{tabular}{c}
\begin{minipage}{0.5\hsize}
\centerline{\includegraphics[width=1.0\linewidth]{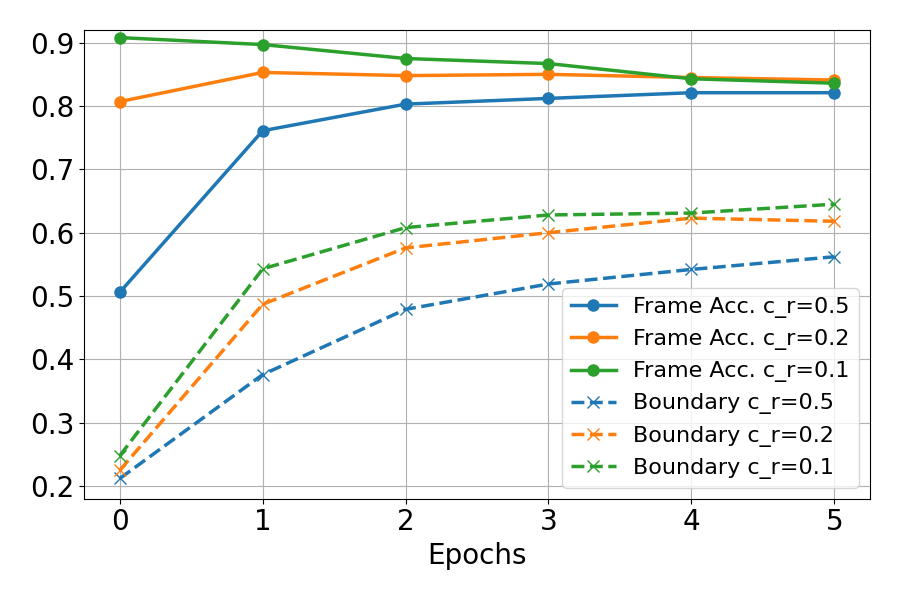}}
\vspace{-1em}
\caption{Accuracy of noisy labels when initialized by corrupted ground-truth labels. Horizontal axis shows elapsed epochs (``0'' denotes initial labels). Vertical axis shows frame accuracy (solid) and boundary score (dashed).}
\label{fig:gplc_corruption}
\end{minipage}
\begin{minipage}{0.5\hsize}
\centerline{\includegraphics[width=0.9\linewidth]{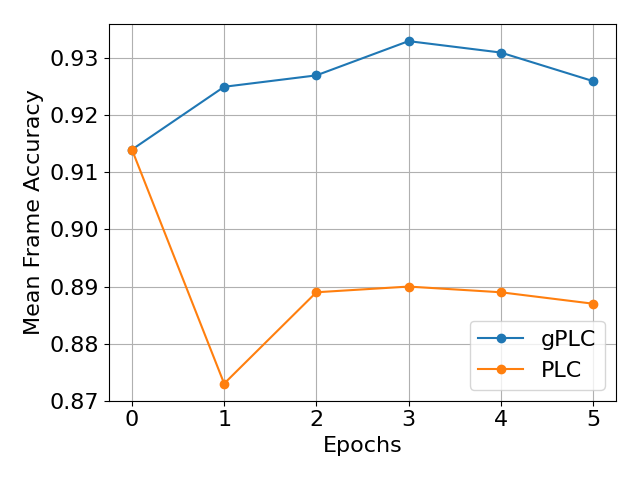}}
\vspace{-1em}
\caption{Accuracy of noisy labels when initialized by motion-based pseudo-labels. Horizontal axis shows elapsed epochs and vertial axis shows mean frame accuracy per track. Note that non-labeled frames are ignored.}
\label{fig:gplc_pl}
\end{minipage}
\end{tabular}
\end{figure}
\vspace{-1em}

To understand the behavior of gPLC, we measured how gPLC corrects labels during training.
We included the validation set into the training data with two patterns of initial labels: (i) randomly corrupted labels from ground truth (with three different corruption ratios $c_r=0.1/0.2/0.5$) (ii) motion-based pseudo-labels.
We trained the full model and measured the accuracy of the labels for every epoch.

First, gPLC succeeded to correct randomly corrupted label even in the case of high corruption ratio of 0.5 (see Figure~{\ref{fig:gplc_corruption}}).
However, in the case of a small corruption ratio of 0.1, gPLC made wrong corrections which  means  that  both  the  noisy  model  and  clean  model got the prediction wrong.
Improved boundary scores showed that gPLC can iteratively suppress inconsistent boundary errors.
In the more realistic case of motion-based pseudo-labels, pseudo-labels were assigned to around 44\% of the total frames, and achieved initial mean frame accuracy of 91.4\% for the labeled frames.
While gPLC reduced the error rate by 20\% PLC wrongly flipped the contact state, which may have harmed the final performance (see Figure~{\ref{fig:gplc_pl}}).
These results indicate that gPLC effectively corrects noisy labels during training.

\section{Conclusion}
We have presented a simple yet effective method of predicting the contact state between hands and objects.
We have introduced a semi-supervised framework of motion-based pseudo-label generation and guided progressive label correction that corrects noisy pseudo-labels guided by a small amount of trusted data.
We have newly collected annotation for evaluation and showed the effectiveness of our framework against several baseline methods.

\section*{Acknowledgements}
This work was supported by JST AIP Acceleration Research Grant Number JPMJCR20U1, JSPS KAKENHI Grant Number JP20H04205 and JP21J11626, Japan.
TY was supported by Masason Foundation.
We are grateful for the insightful suggestions and feedback from the anonymous reviewers.

\newpage
\appendix
\setcounter{section}{1}
\subsection{Details on Pseudo-Label Generation}
\paragraph*{Preprocessing}
We extracted frames from videos by either 30 or 25 fps, half of the original frame rate.
We processed all the frames in the resolution of $854\times480$.

\paragraph*{Hyperparameters}
In the implementation, we used different $\sigma$ and motion thresholds for contact detection and no-contact detection.
We summarize the hyperparameters used in Table~\ref{tab:pl_hyperparameters}.
$d_h$ and $d_o$ denote the average motion direction of the hand region and object region.
We tuned the hyperparameter using the validation set.

For pseudo-label extension, we tracked at most 100 points for each hand and object to track the distance between hands and objects.

\begin{table}[t]
    \centering
    \begin{tabular}{lrrrrr}
    \hline
        & $\sigma$ & hand & object & background & motion direction \\ \hline
        Contact & 2.0 & $h_r \ge 0.7$ & $o_r < 0.05$ & $b_r < 0.2$ & $sim(d_h, d_o) > 0.5$\\
        No-contact & 1.0 & $h_r \ge 0.7$ & $o_r \ge 0.2$ & $b_r < 0.2$ & $sim(d_h, d_o) < 0.0$ \\
        \hline
    \end{tabular}
    \caption{Hyperparameters used for pseudo-label generation.}
    \label{tab:pl_hyperparameters}
\end{table}

\paragraph*{Additional examples}
\begin{figure}[t]
\begin{minipage}[t]{1\hsize}
\centerline{\includegraphics[width=0.9\linewidth]{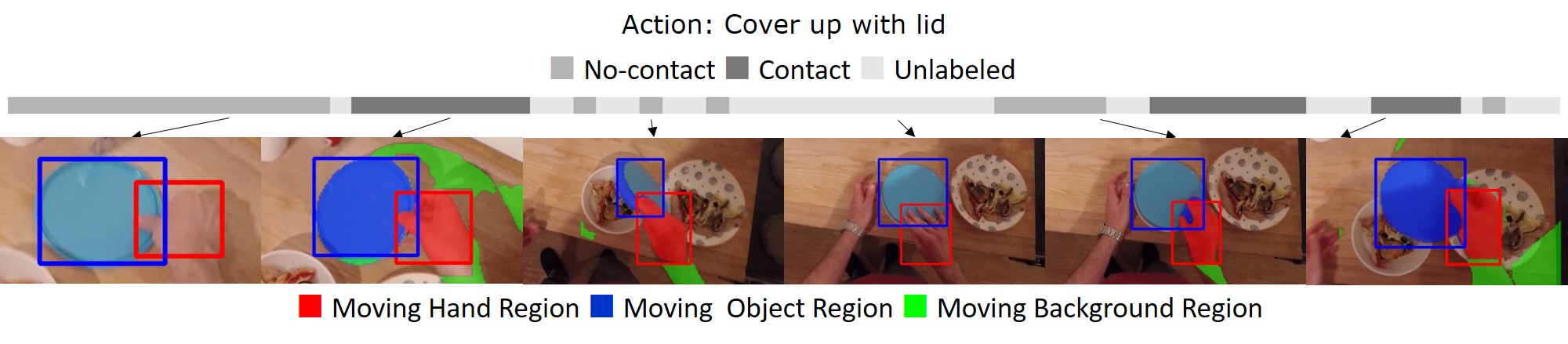}}
\end{minipage}
\begin{minipage}[t]{1\hsize}
\centerline{\includegraphics[width=0.9\linewidth]{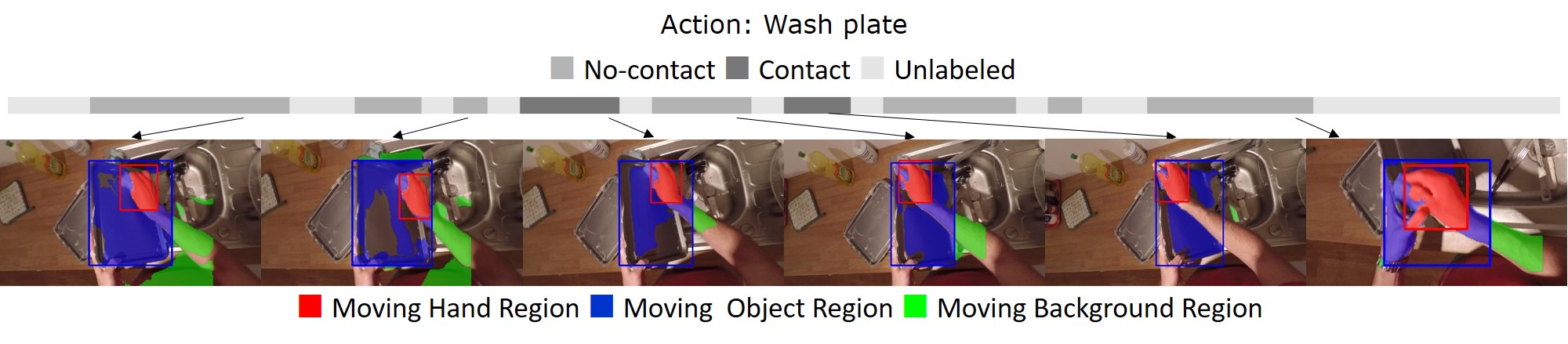}}
\end{minipage}
\begin{minipage}[t]{1\hsize}
\centerline{\includegraphics[width=0.9\linewidth]{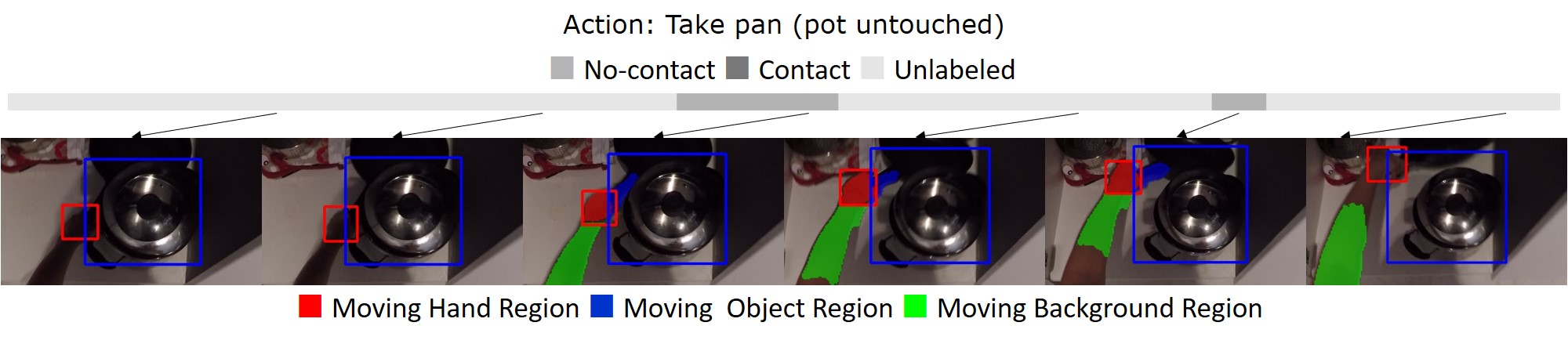}}
\end{minipage}
\begin{minipage}[t]{1\hsize}
\centerline{\includegraphics[width=0.9\linewidth]{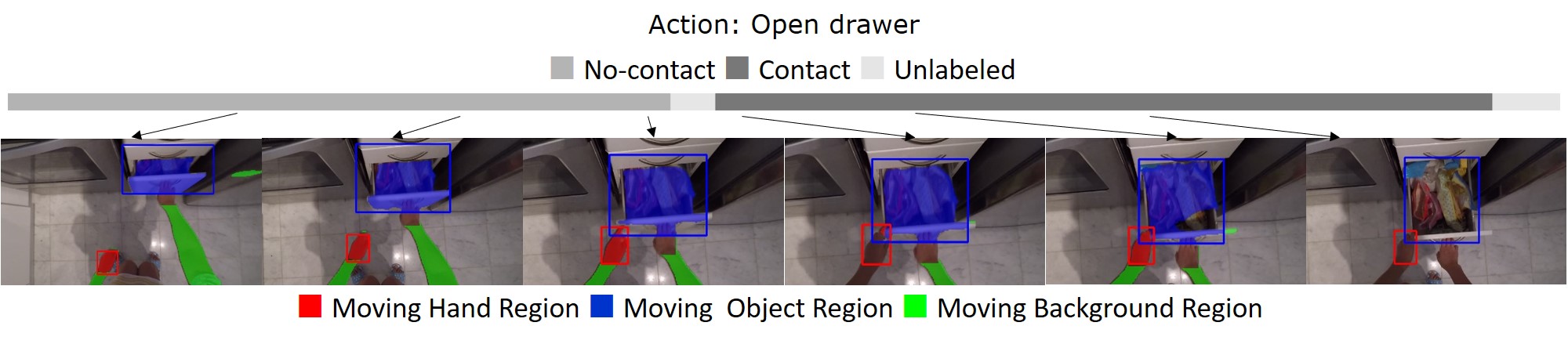}}
\end{minipage}
\begin{minipage}[t]{1\hsize}
\centerline{\includegraphics[width=0.9\linewidth]{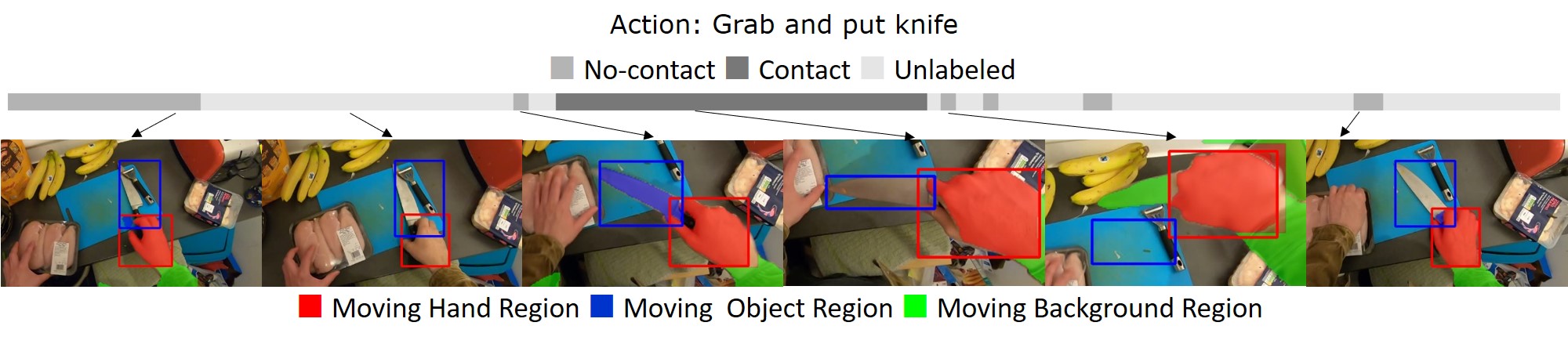}}
\end{minipage}
\caption{{\bf Additional examples of generated pseudo-labels:} (Top) Gray and dark gray bar indicates no-contact/contact labels otherwise no labels assigned. (Bottom) Representative frames. Red, blue, and green regions denote moving hand, object, and background regions, respectively. Note that in few tracking errors are included in these tracks (\eg, rightmost frame in second example). Refer to video visualization for detail.}
\label{fig:add_pseudo_labels}
\end{figure}

Figure~\ref{fig:add_pseudo_labels} shows additional results on pseudo-label generation.
As seen in the figures, our procedure assigns reliable pseudo-labels in various types of interactions.
However, few tracking errors are included (\eg, rightmost frame in second example) and the label assignment is not perfect, suggesting the needs of correction.

\subsection{Model Details}
\paragraph*{Network architecture}
\begin{figure}[t]
\centerline{\includegraphics[width=0.9\linewidth]{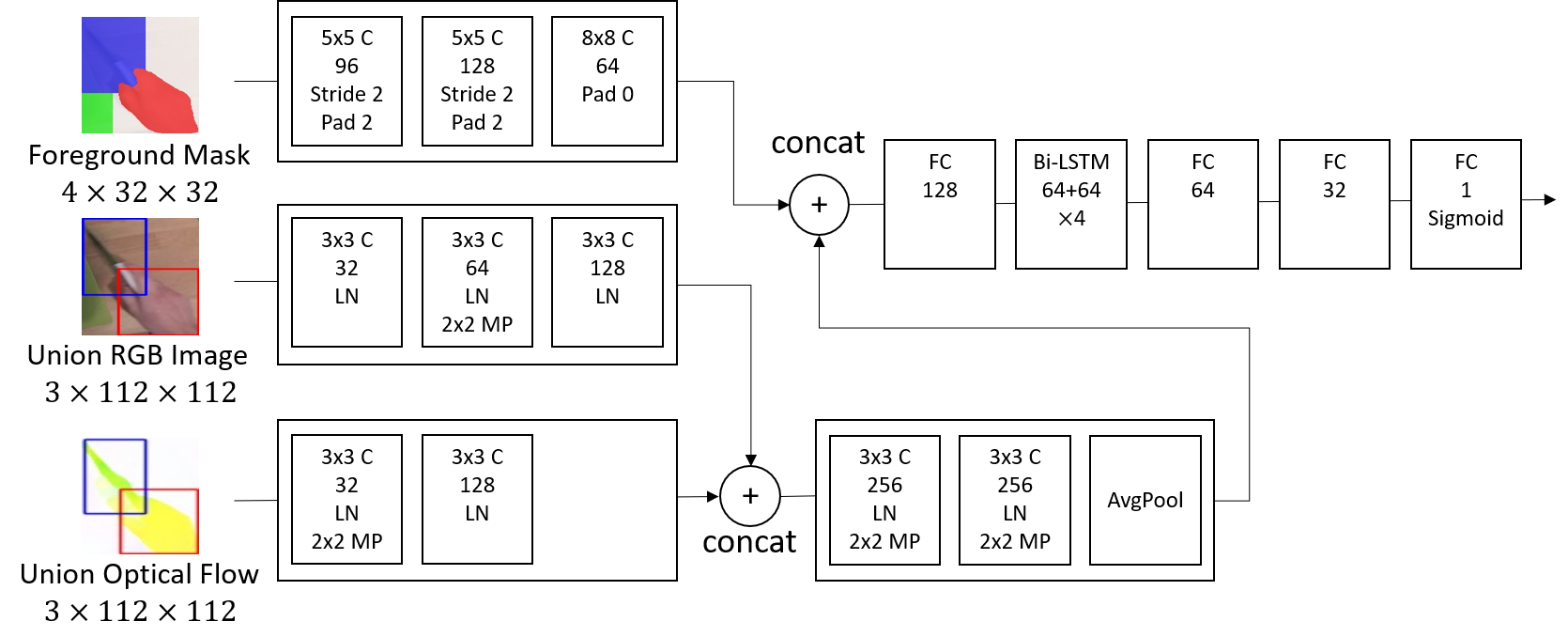}}
\caption{Detailed network architecture of contact prediction model (per frame). C denotes convolutional layer with filter size and number of channels, followed by ReLU layer. MP denotes max pooling with filter size. LN denotes layer normalization layer. FC denotes fully-connected layer with number of units, followed by ReLU layer (except last layer).}
\label{fig:method_detail}
\end{figure}

Figure~\ref{fig:method_detail} shows the architecture of the proposed contact prediction model.
The input frames are passed one by one and temporal dependencies will be captured at the bidirectional LSTM layers.

\paragraph*{Training}
During training, if the length of the hand-object track is long, we randomly cropped the track at a maximum length of 105 to fit the GPU memory.

\paragraph*{Baseline models}
In {\bf ContactHands} and {\bf Shan-\textasteriskcentered}, we used the pre-trained model provided by the authors.
We used the combined model in the former and the model trained on the 100DOH dataset and egocentric datasets for the latter.
In those baseline models, we link between the predicted hand instance mask and ground truth hand instance mask if IoU is above 0.5.
In {\bf Shan-Bbox}, we predicted as a contact if IoU between a predicted object bounding box and a ground truth bounding box is larger than 0.5.
{\bf Shan-Full} combines {\bf Shan-Bbox} and {\bf Shan-Contact} based on the following rules: (i) If IoU between input hand instance mask and input object bounding box is zero, predicts as a no-contact; (ii) If {\bf Shan-Bbox} predicts as a no-contact, follow its prediction; (iii) Otherwise, use predictions produced by {\bf Shan-Contact}.
We observed improved performance by combining predictions based on object-in-contact detection and contact state prediction.

\subsection{Dataset Details}

\paragraph*{Pseudo-labels}
We used 96,000 tracks (9 million frames) with bounding boxes and pseudo-labels for training.
Pseudo-labels were assigned for 37.3\% of the total frames.

\paragraph*{Trusted labels}
We annotated 67,064 frames of 1,200 tracks.
We did not annotate the instance masks of the hands since the segmentation network described in Section~3.1 produced reliable results.
The average length of the track was 56 frames.
To evaluate whether the model can distinguish touched and untouched objects, we included tracks stably in contact and untouched tracks.
The number of tracks that were in constant contact was 284, the number of tracks with mixed contact states was 670, and the number of tracks that were not in contact was 246.

\subsection{Additional Experimental Results}
\paragraph*{Effect of pseudo-label set size.}
Table~\ref{tab:ablation_3} shows ablation results on changing the noisy dataset size.
We sampled 1\%, 5\%, and 25\% of the full noisy dataset and trained by the proposed gPLC algorithm.
The result supports the fact that using large-scale data with pseudo-labels helps generalization.

\begin{table}[t]
\begin{center}
\scalebox{.8}{
\begin{tabular}{lrrrrr}
\hline
Amount of pseudo-labels & Frame Acc. & Boundary Score & Peripheral Acc. & Edit Score & Correct Ratio\\ \hline
0\% (supervised-train) & 0.770 & 0.563 & 0.649 & 0.718 & 0.394 \\ %
1\% & 0.784 & 0.595 & 0.728 & 0.729 & 0.397 \\ %
5\% & 0.803 & 0.620 & 0.737 & 0.747 & 0.427 \\ %
25\% & 0.818 & 0.651 & 0.725 & 0.772 & 0.467 \\ %
100\% (proposed) & 0.836 & 0.681 & 0.730 & 0.793 & 0.519 \\ \hline %
\end{tabular}
}
\caption{Ablations on noisy dataset size.}
\label{tab:ablation_3}
\end{center}
\end{table}

\begin{figure}[t]
\begin{tabular}{c}
    \begin{minipage}[t]{1\hsize}
    \centerline{\includegraphics[width=0.95\linewidth]{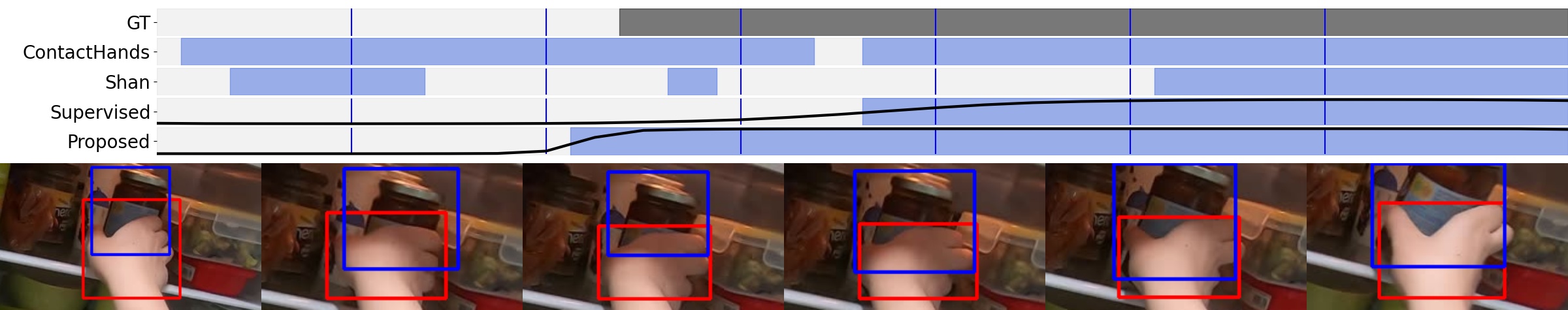}} %
    \end{minipage} \\ 
    \begin{minipage}[t]{1\hsize}
    \centerline{\includegraphics[width=0.95\linewidth]{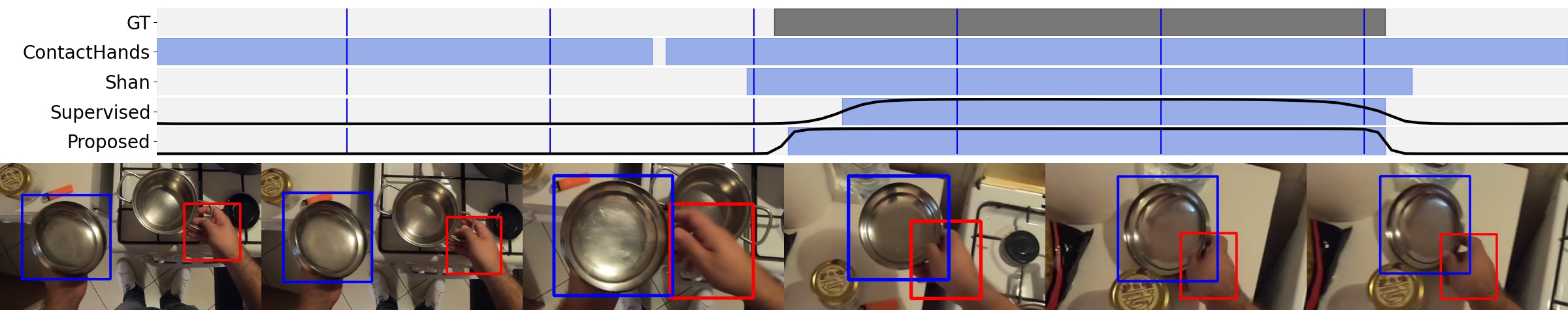}} %
    \end{minipage} \\
    \begin{minipage}[t]{1\hsize}
    \centerline{\includegraphics[width=0.95\linewidth]{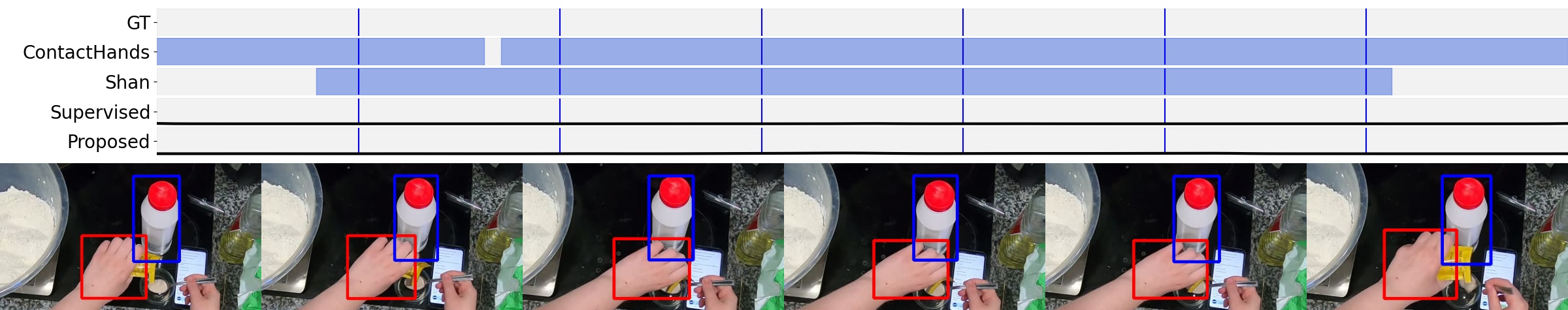}} %
    \end{minipage} \\
    \begin{minipage}[t]{1\hsize}
    \centerline{\includegraphics[width=0.95\linewidth]{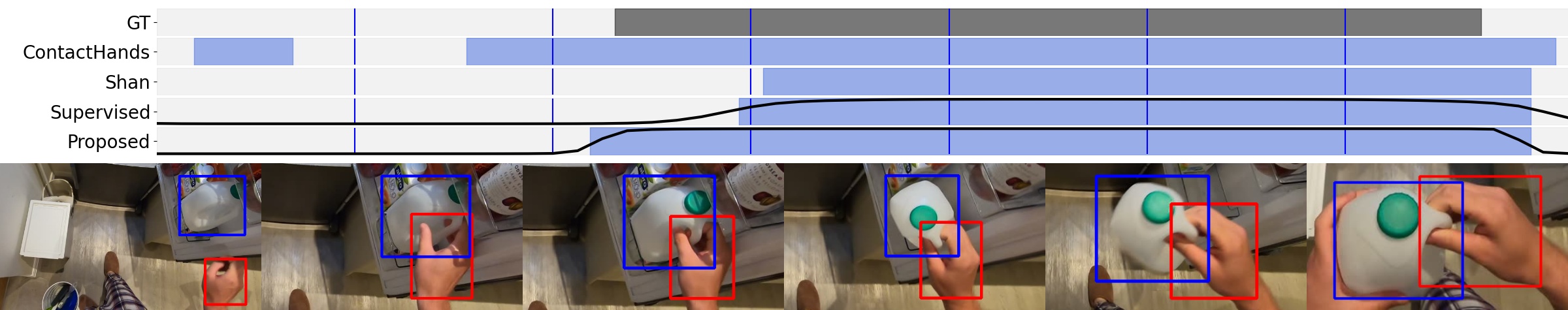}} %
    \end{minipage} \\
    \begin{minipage}[t]{1\hsize}
    \centerline{\includegraphics[width=0.95\linewidth]{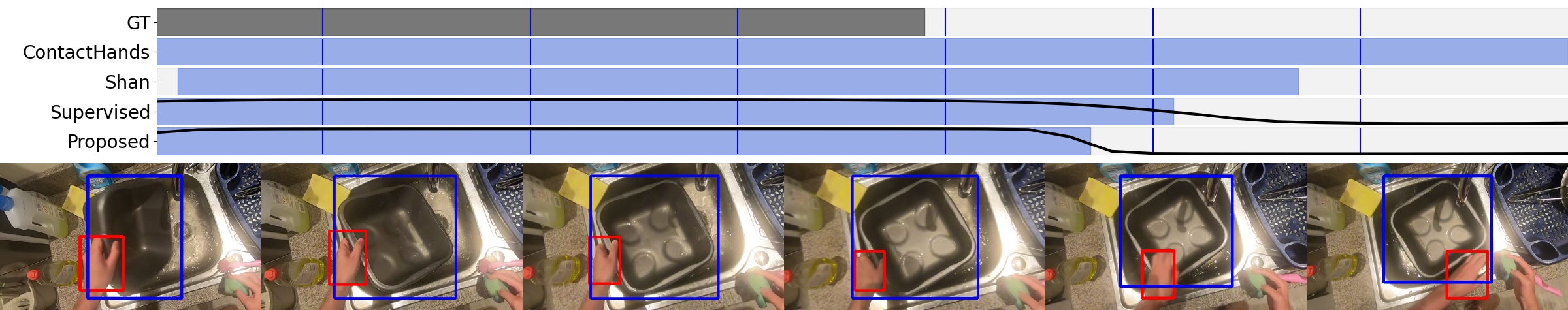}} %
    \end{minipage} \\
    \begin{minipage}[t]{1\hsize}
    \centerline{\includegraphics[width=0.95\linewidth]{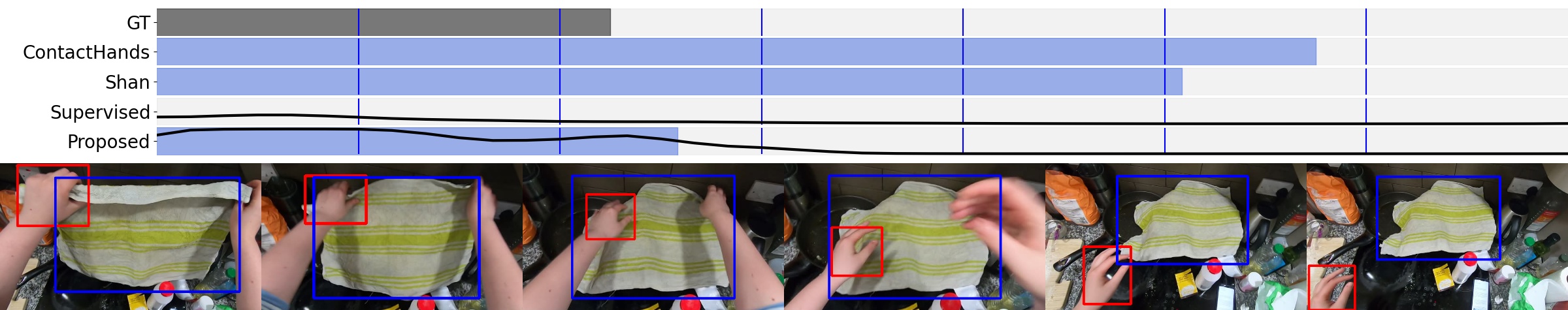}} %
    \end{minipage} \\
\end{tabular}
\caption{Additional qualitative examples. Our model better predicts correct contact state change point and can avoid false positives in difficult cases of image-level overlap between hand and objects.  Refer to video visualization for detail.}
\label{fig:add_qual_result}
\end{figure}

\paragraph*{Additional qualitative examples}
\begin{figure}[t]
\begin{tabular}{c}
    \begin{minipage}[t]{1\hsize}
    \centerline{\includegraphics[width=0.95\linewidth]{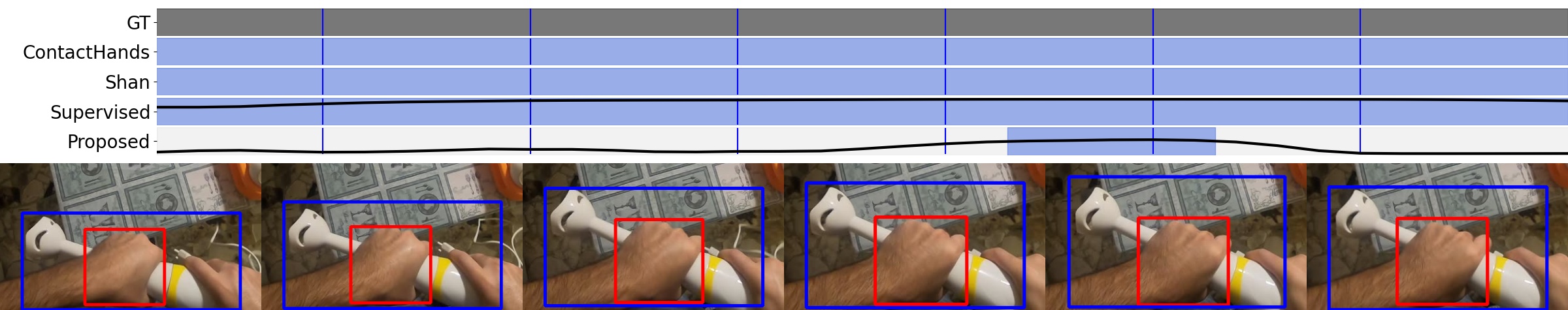}} %
    \end{minipage} \\ 
    \begin{minipage}[t]{1\hsize}
    \centerline{\includegraphics[width=0.95\linewidth]{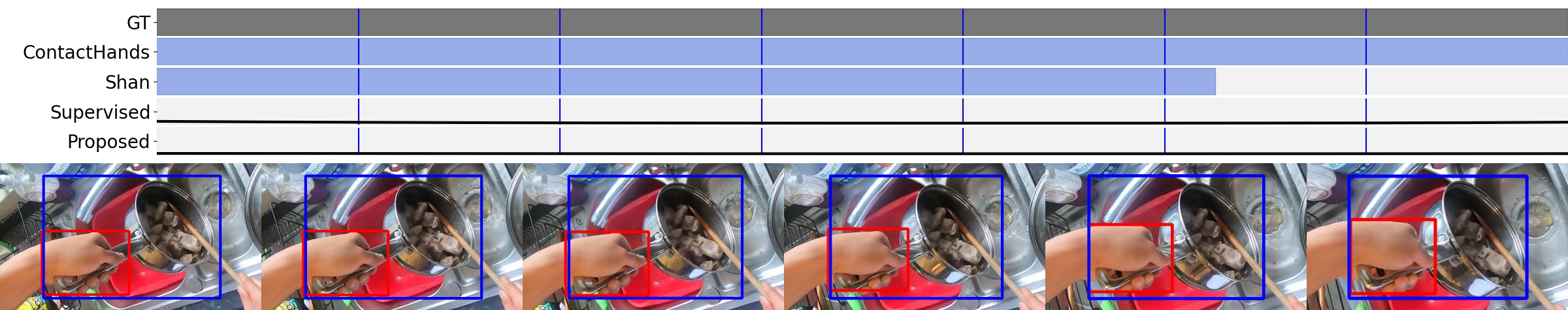}} %
    \end{minipage} \\
    \begin{minipage}[t]{1\hsize}
    \centerline{\includegraphics[width=0.95\linewidth]{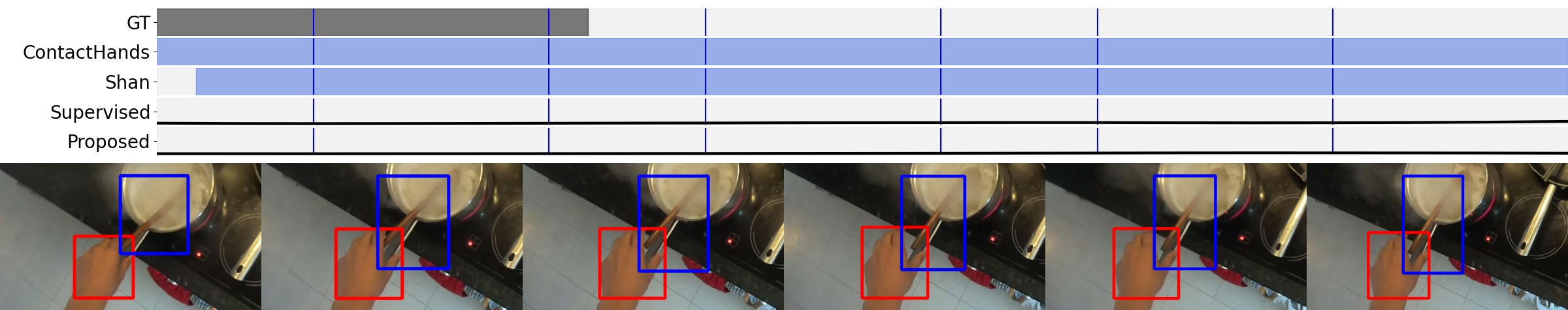}} %
    \end{minipage} \\
\end{tabular}
\caption{Additional failure examples.}
\label{fig:add_failures}
\end{figure}
Figure~\ref{fig:add_qual_result} shows additional examples on the contact prediction result.
Our model focused on motion rather than spatial overlap to infer the contact states.
The performance was boosted at novel objects thanks to large-scale pseudo-label training.

We also show additional failure cases in Figure~\ref{fig:add_failures}.
We observed failures due to unfamiliar grasps (top), no movement (middle), subtle hand movement (bottom).
Since we focused on the holistic foreground motion of hands and objects, it was difficult to predict contacts in fine-grained manipulation.

\paragraph*{Effect of input modality}
\begin{figure}[t]
\begin{tabular}{c}
    \begin{minipage}[t]{1\hsize}
    \centerline{\includegraphics[width=0.95\linewidth]{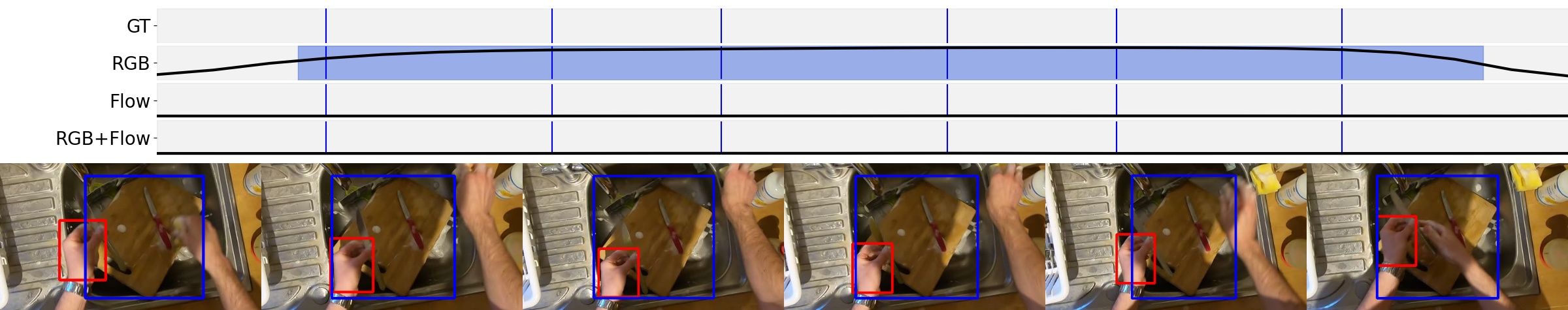}} %
    \end{minipage} \\ 
    \begin{minipage}[t]{1\hsize}
    \centerline{\includegraphics[width=0.95\linewidth]{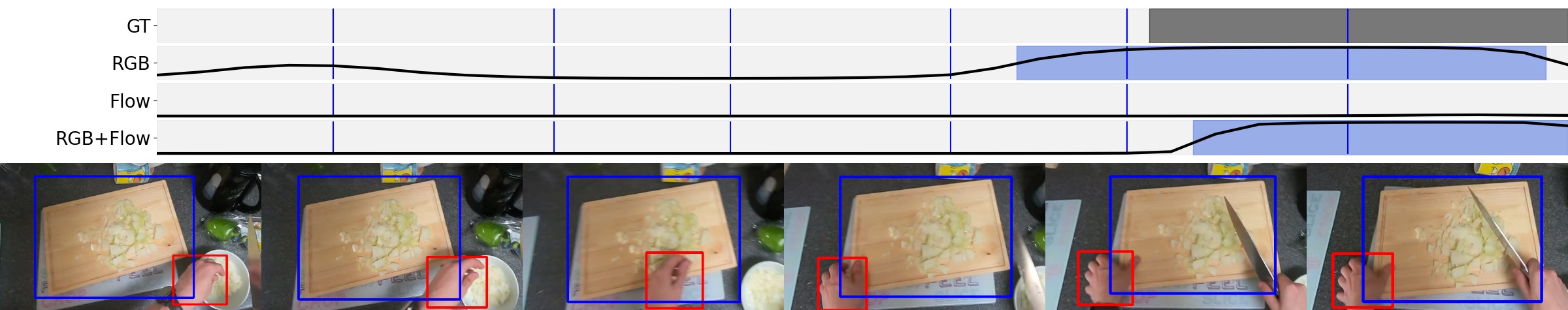}} %
    \end{minipage} \\
    \begin{minipage}[t]{1\hsize}
    \centerline{\includegraphics[width=0.95\linewidth]{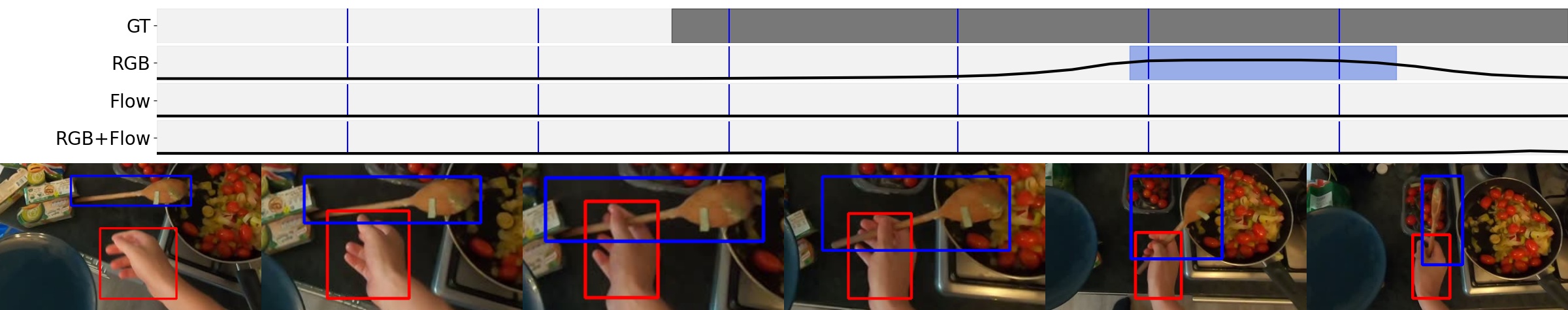}} %
    \end{minipage}
\end{tabular}
\caption{Qualitative results on input modalities.}
\label{fig:qual_modality}
\end{figure}

Figure~\ref{fig:qual_modality} shows the prediction examples on different input modalities.
In general, the model trained by RGB input solely tended to make uncertain predictions near boundaries (see Figure~\ref{fig:qual_modality} top).
Also shown in the lower boundary score, this result indicates distinguishing a contact state is difficult from a single image.
The model trained by flow input solely generally behaves similar to the proposed model.
However, the difference appears when there is no motion in the scene.
If there is no or subtle motion in the scene, the flow model has no clue except temporal context to predict contacts.
In such cases, RGB images will be the only clue (see Figure~\ref{fig:qual_modality} middle).
Motion information contributed to accurate prediction in most cases but sometimes failed when complex motion patterns are observed (see Figure~\ref{fig:qual_modality} bottom).

\end{document}